\documentclass[10pt,twocolumn,letterpaper]{article}

\usepackage[pagenumbers]{cvpr} 

\usepackage[dvipsnames]{xcolor}

\usepackage{graphicx}
\usepackage{amsmath}
\usepackage{amssymb}
\usepackage{booktabs}
\usepackage{cuted}
\usepackage{caption}
\usepackage{cuted}
\usepackage{float}
\usepackage{graphicx}
\usepackage{pifont}
\usepackage{xcolor}
\usepackage{listings}
\usepackage{bm}
\usepackage[T1]{fontenc}

\usepackage{ragged2e}
\usepackage{booktabs}
\usepackage{color}
\usepackage{multirow}
\usepackage{bm}
\usepackage{bbm}
\usepackage{amsmath}
\usepackage{algorithm, algpseudocode}
\usepackage{amsfonts}
\usepackage{enumerate}

\usepackage{cuted}
\usepackage{capt-of}

\newcommand{\figref}[1]{Fig. \ref{#1}}
\newcommand{\tabref}[1]{Tab. \ref{#1}}

\newcommand{\cmark}{\ding{51}}%
\newcommand{\xmark}{\ding{55}}%

\newcommand{\projpage}{https://magic-research.github.io/dream-talk/}

\definecolor{cvprblue}{rgb}{0.21,0.49,0.74}
\usepackage[pagebackref,breaklinks,colorlinks,citecolor=cvprblue]{hyperref}

\def\paperID{} 
\def\confName{CVPR}
\def\confYear{2024}

\title{DREAM-Talk: \textcolor{Red}{D}iffusion-based \textcolor{Red}{R}ealistic \textcolor{Red}{E}motional \textcolor{Red}{A}udio-driven \textcolor{Red}{M}ethod for Single Image \textcolor{Red}{Talk}ing Face Generation}

\author{Chenxu Zhang\textsuperscript{1*}, Chao Wang\textsuperscript{1}\thanks{Equal contribution} , Jianfeng Zhang\textsuperscript{1}, Hongyi Xu\textsuperscript{1}, Guoxian Song\textsuperscript{1}, You Xie\textsuperscript{1}, \\
Linjie Luo\textsuperscript{1}, Yapeng Tian\textsuperscript{2}, Xiaohu Guo\textsuperscript{2}, Jiashi Feng\textsuperscript{1}\\
\textsuperscript{1}ByteDance Inc.\qquad\textsuperscript{2}The University of Texas at Dallas\\
}

\begin{document}
\maketitle

\begin{strip}
    \centering
    \vspace{-4em}
    \centering
    \includegraphics[width=\textwidth]{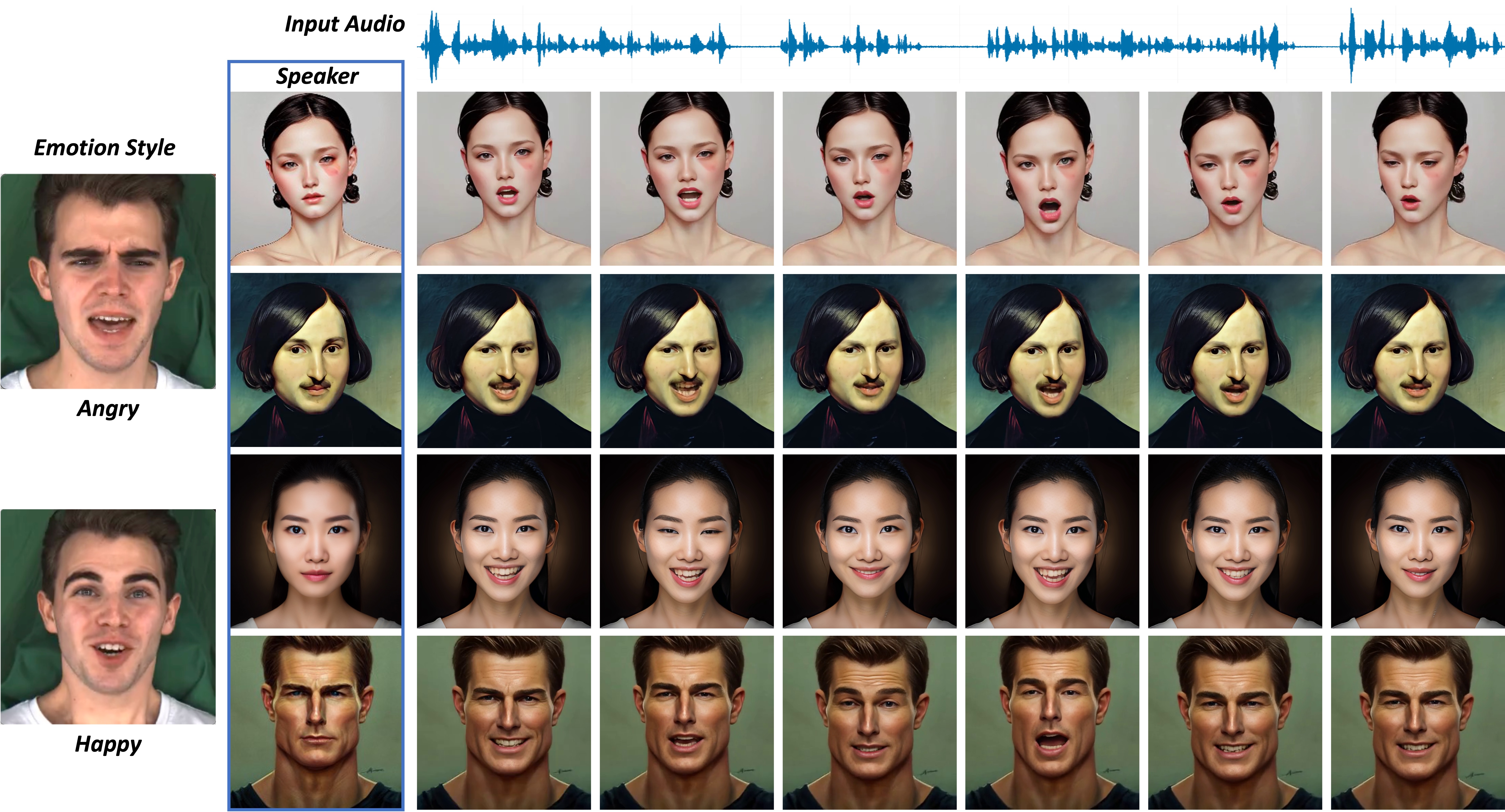}
    \vspace{-2em}
    \captionof{figure}{\textbf{DREAM-Talk} takes as input a driving audio sequence, a given portrait image, and an example of emotion style (a clip of an emotional talking face), and generates a photorealistic, lip-synchronized talking face video that features high-quality emotional expressions. The results include both real human images and images generated by AIGC. Please refer to our \href{\projpage}{Project Page} for more results.}
    \label{fig:teaser}
    \vspace{-1mm}
\end{strip}


\begin{abstract}
The generation of emotional talking faces from a single portrait image remains a significant challenge. 
The simultaneous achievement of expressive emotional talking and accurate lip-sync is particularly difficult, as expressiveness is often compromised for the accuracy of lip-sync. As widely adopted by many prior works, the LSTM network often fails to capture the subtleties and variations of emotional expressions.
To address these challenges, we introduce DREAM-Talk, a two-stage diffusion-based audio-driven framework, tailored for generating diverse expressions and accurate lip-sync concurrently.
In the first stage, we propose EmoDiff, a novel diffusion module that generates diverse highly dynamic emotional expressions and head poses in accordance with the audio and the referenced emotion style.
Given the strong correlation between lip motion and audio, we then refine the dynamics with enhanced lip-sync accuracy using audio features and emotion style. 
To this end, we deploy a video-to-video rendering module to transfer the 
expressions and lip motions from our proxy 3D avatar to an arbitrary portrait. Both quantitatively and qualitatively, DREAM-Talk outperforms state-of-the-art methods in terms of expressiveness, lip-sync accuracy and perceptual quality. 

\end{abstract}

\section{Introduction}
\label{sec:intro}

The field of talking face generation has seen significant advancements in recent years, becoming a key area of research with a wide range of applications, including video conferencing, virtual assistants, and entertainment, among others. 
Recently, researchers have commenced incorporating emotion-conditioned facial expressions into talking face generation~\cite{ji2021audio,ji2022eamm,gururani2022spacex}, leveraging emotional annotations in talking video datasets like MEAD~\cite{wang2020mead}. However, none of these methods have yet succeeded in generating expressive and lifelike expressions in their talking faces.

Several major challenges exist in current emotional talking face generation methods.
Firstly, it is difficult to achieve \emph{expressive emotion} and \emph{accurate lip-sync} simultaneously. Emotional expressions in datasets like MEAD~\cite{wang2020mead} show significant exaggeration in the movements of eyebrows, eye-blinking, and mouth shapes. Nonetheless, the sentences and audio content used in these datasets lack sufficient length to effectively train a precise lip-sync model.
To address this issue, 
SPACE~\cite{gururani2022spacex} employed two supplementary non-emotional datasets, namely VoxCeleb2~\cite{chung18b_interspeech} and RAVDESS~\cite{Livingstone18RAVDESS}, alongside MEAD, to train their model.  
However, integrating non-emotional and emotional datasets results in synthesized emotional expressions that may lack the desired level of expressiveness and dynamism. EAMM~\cite{ji2022eamm} tackles this problem by integrating two modules: one dedicated to learning non-emotional audio-driven face synthesis and another focused on capturing emotional displacements of expressions. To prevent emotional displacements from distorting lip-sync, it augmented the training data by obscuring the mouth region of the speakers. Unfortunately, employing mouth-covering data augmentation compromises the expressiveness of mouth shapes during emotional speech.

Secondly, modeling the subtleties and variations of emotional expressions is challenging. Emotional expressions involve the activation of numerous facial muscles and exhibit significant diversity throughout a speech. Existing methods~\cite{ji2021audio,ji2022eamm,ma2023styletalk,gururani2022spacex} typically utilize LSTM or CNN networks as generators to transform audio into facial representations. While these models are adequate for capturing the movements of the mouth and lips during regular speech, they face challenges when it comes to faithfully portraying the nuances and variations of emotional expressions. Consequently, their generated emotional depictions often appear bland and artificial. For a detailed comparison of these approaches, please refer to the supplementary video.

To overcome these challenges,  we propose a diffusion-based realistic emotional audio-driven method for talking face generation (DREAM-Talk) from a single portrait image. 
At its core is a carefully designed two-stage pipeline, which achieves both expressive emotion and precise lip-sync, as shown in Fig.~\ref{fig:framework}.
The first stage, 
\emph{EmoDiff Module}, is tailored to capture the dynamic nature of emotional expressions. Specifically, we designed an emotion-conditioned diffusion model to transform input audio to the facial expressions of the ARKit model~\cite{liu2022beat}.
The second stage, \emph{Lip Refinement}, focuses on ensuring the precision of lip-sync in the generated talking faces. 
To enhance the synchronization of mouth movements with audio signals while preserving the richness of emotional expressions, we have developed a novel lip refinement network to re-optimize the parameters of the mouth based on audio signals and specific emotional styles.
Unlike traditional face model~\cite{li2017learning} where mouth parameters are integrated with other facial parameters, using 3D ARKit model enables explicitly optimizing lip motion, ensuring that the intensity of other facial expressions remains unaffected. This design choice in our lip refinement network guarantees that the expressiveness of emotions is not compromised by lip-sync refinement, offering a more targeted and effective approach for emotion-rich facial animation.

The sequential two-stage process employed in DREAM-Talk effectively addresses the challenges mentioned earlier, allowing for the simultaneous achievement of expressive emotions and precise lip-sync in the generated talking faces. Our experimental results convincingly showcase its exceptional ability to model the intricacies and variations of emotional expressions from the input audio. This includes realistically capturing emotional movements in areas such as eyebrows, eye blinks, mouth shapes, and beyond. Specifically, our diffusion model adeptly captures high-frequency facial details, while lip refinement further elevates the precision of mouth motion.
The contributions of this paper can be summarized as follows:

\begin{figure*}[t]
\begin{center}
   \includegraphics[width=\linewidth]{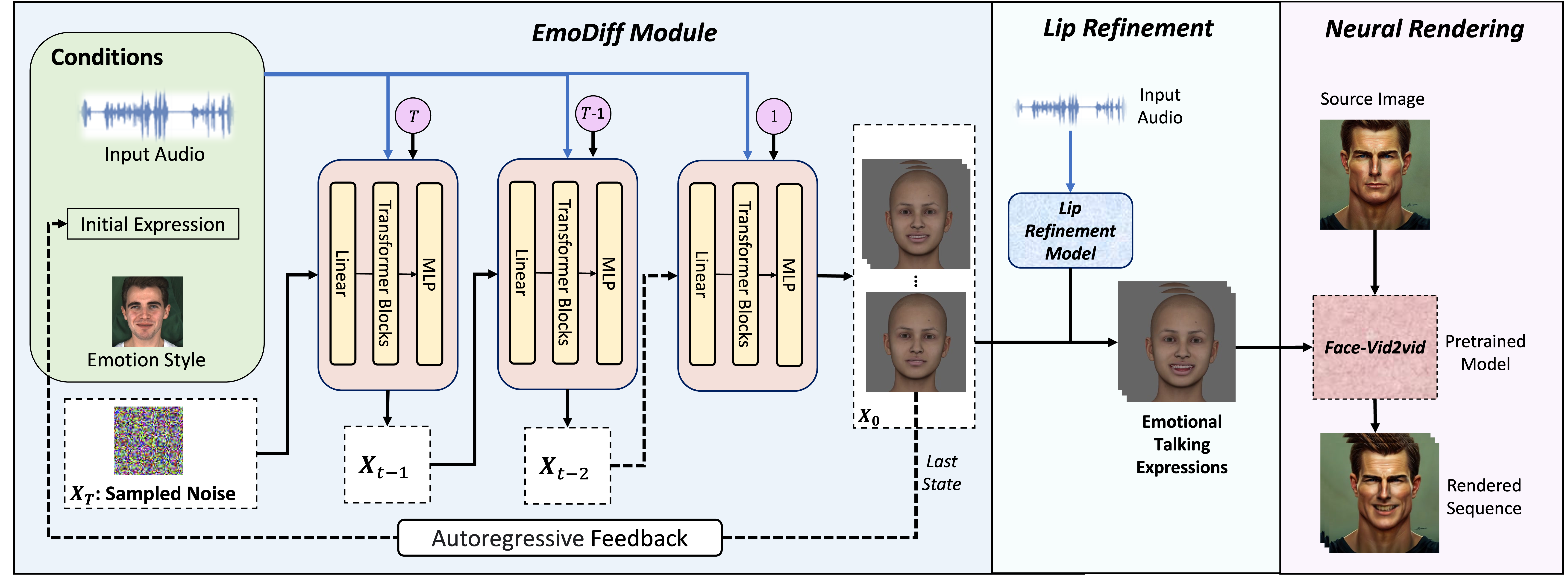}
\end{center}
   \caption{
   Pipeline of our DREAM-Talk framework. Starting with the input audio, initial state, and emotion style as conditions, we first employ EmoDiff for learning to denoise 3D expressions over time, utilizing a transformer-based architecture for sequence modeling. The initial state corresponds to the expression in the first frame, and the emotion style is defined by a randomly selected expression clip, independent of the input audio. Then, utilizing the conditioned audio and emotional expressions, the lip refinement model further optimizes the mouth without altering the intensity of emotions. This is followed by producing corresponding 3D rendering faces on a blendshape rig. Finally, we employ a fine-tuned Face-Vid2Vid model~\cite{wang2021facevid2vid} to generate emotional talking videos.
   } 
\vspace{-2mm}
\label{fig:framework}
\end{figure*}
\begin{itemize}
\item A novel two-stage generative framework DREAM-Talk that achieves expressive emotional talking faces with audio-synchronized lip motion. 

\item  A diffusion module EmoDiff that is capable of generating diverse highly-dynamic emotional expressions and head poses in accordance with the audio and the referenced emotion style. 

\item An emotional ARKit dataset with precise isolation of mouth parameters from other facial attributes, well tailored for the task of lip motion refinement. 
\end{itemize}

\section{Related Work}
\label{sec:related_work}

\noindent
\textbf{Audio-driven talking face}. The task of generating a talking face driven by audio involves producing a realistic and cohesive video of a person's speaking face, utilizing audio input and, occasionally, an image or video of the speaker. Early efforts by Taylor et al. \cite{taylor2017deep} focused on converting audio sequences into phoneme sequences to generate adaptable talking avatars for multiple languages. \cite{suwajanakorn2017synthesizing} curated a dataset of Obama videos and introduced an end-to-end framework to synthesize corresponding talking faces with arbitrary voices. \cite{chung2017you} and \cite{zhou2020makelttalk} pioneered the generation of talking face videos, requiring only a single facial and audio sequence. ATVGnet \cite{chen2019hierarchical} and \cite{zhong2023identitypreserving} proposed two-stage talking face synthesis methods guided by landmarks. They initially generated landmarks from a single identity image and an audio sequence, which were then combined with the identity image for the second stage of talking face synthesis. \cite{meshry2021learned} decomposed talking head synthesis into spatial and style components, demonstrating improved performance in few-shot novel view synthesis. FACIAL \cite{zhang2021facial} and \cite{song2022everybody} used audio to regress parameters in 3D face models, resulting in more realistic synthesis. 
To enhance video quality, AD-NeRF \cite{guo2021ad} and Geneface \cite{ye2023geneface} employed an audio-driven neural radiance fields (NeRF) model to synthesize high-quality talking-head videos from audio input, surpassing existing GAN-based methods. 
SadTalker \cite{zhang2023sadtalker} adeptly generates emotive speech content by mapping audio inputs to 3DMM motion coefficients, but challenges remain in achieving both realistic expression and accurate lip movement.
On the other hand, \cite{stypulkowski2023diffused} and DiffTalk \cite{shen2023difftalk} applied diffusion models to avoid challenges of GAN-based methods, such as training difficulties, mode collapse, and facial distortion.  
However, 
those methods require extra motion sequence of the target individual to guide the video generation and avoid unnatural-looking motions.
Moreover, utilizing diffusion model as foundational framework, DiffTalk also encountered challenges in maintaining temporal coherence within the mouth region. On the other hand, the works discussed above still lack emotional information guidance, leading to monotonous expressions in generated talking faces.

\vspace{2mm}
\noindent
\textbf{Emotional audio-driven talking face}.
In recent research endeavors, there has been a growing focus on the development of emotionally expressive talking faces \cite{xu2023high,ji2022eamm, tan2023emmn,gan2023efficient} or emotional talking mesh \cite{peng2023emotalk}. ExprGAN \cite{ding2018exprgan} introduced an expression control module that enables the synthesis of faces with diverse expressions, allowing for continuous adjustment of expression intensity. \cite{eskimez2021speech} presented a neural network system conditioned on categorical emotions, providing direct and flexible control over visual emotion expression. MEAD \cite{wang2020mead} enhanced the vividness of talking faces by curating the MEAD dataset, offering a baseline for emotional talking face generation.
\cite{ji2021audio} proposed a groundbreaking method for emotional control in video-based talking face generation, incorporating a component for distilling content-agnostic emotion features. Addressing the challenge of the timbre gap, \cite{li2021write} introduced a framework for talking-head synthesis that generates facial expressions and corresponding head animations from textual inputs. 
EAMM \cite{ji2022eamm} and EMMN \cite{tan2023emmn} both utilized 2D keypoints displacement to synthesize the final emotional video, which can degrade the quality of generation.
\cite{liang2022expressive} presented a method for generating expressive talking heads with meticulous control over mouth shape, head pose, and emotional expression. \cite{sinha2022emotion} proposed an optical flow-guided texture generation network capable of rendering emotional talking face animations from a single image, regardless of the initial neutral emotion. SPACE \cite{gururani2022spacex} introduced a method decomposing the task into facial landmark prediction and emotion conditioning, resulting in talking face videos with elevated resolution and fine-grained controllability. 
In our work, we employ a diffusion model to predict expression sequences, yielding more expressive outcomes.
\section{Method}
\subsection{Preliminaries}
\label{sec:preliminaries}

Contrary to 2D landmark-based methods, which are susceptible to head pose variations and often face challenges in maintaining consistent facial shape representation~\cite{zhou2020makelttalk, ji2022eamm}, 3D modeling techniques offer shape-invariant information, thereby facilitating more realistic renderings that align with the actual three-dimensional structure of human faces. Traditional 3D models, such as 3D Morphable Models (3DMM) or FLAME, predominantly utilize Principal Component Analysis (PCA) to encapsulate facial features. While these parameters provide control over general facial appearance, they fall short in isolating specific facial attributes, such as eye blinking or lip movements. Given our objective to enhance the mouth region while concurrently preserving the expressiveness of other facial features, we have elected to employ ARKit blendshapes. This technology distinctly separates mouth-related parameters from other facial elements, thus enabling targeted optimization. The ARKit facial model comprises 52 distinct parameters, each representing unique facial features. It utilizes blendshapes based on the Facial Action Coding System (FACS), allowing each facial expression to activate specific facial regions (e.g., mouth area, eyes, eyebrows) independently and in a manner consistent with human facial anatomy \cite{liu2022beat}. This approach offers precise control over and optimization of various facial attributes, rendering it particularly well-suited for our specialized optimization requirements.

Subsequently, we conduct a comprehensive analysis of ARKit parameters on each frame within the MEAD emotion dataset, thereby extracting corresponding parameters. This process facilitates the creation of an ARKit-specific facial dataset, meticulously tailored to align with the emotional nuances of the MEAD dataset. To the best of our knowledge, we are the first to develop an emotion dataset that features fully disentangled 3D facial parameters. Such a development significantly amplifies the practical utility and applicability of emotion-based datasets in the field.

\begin{figure}[t]
\begin{center}
   \includegraphics[width=\linewidth]{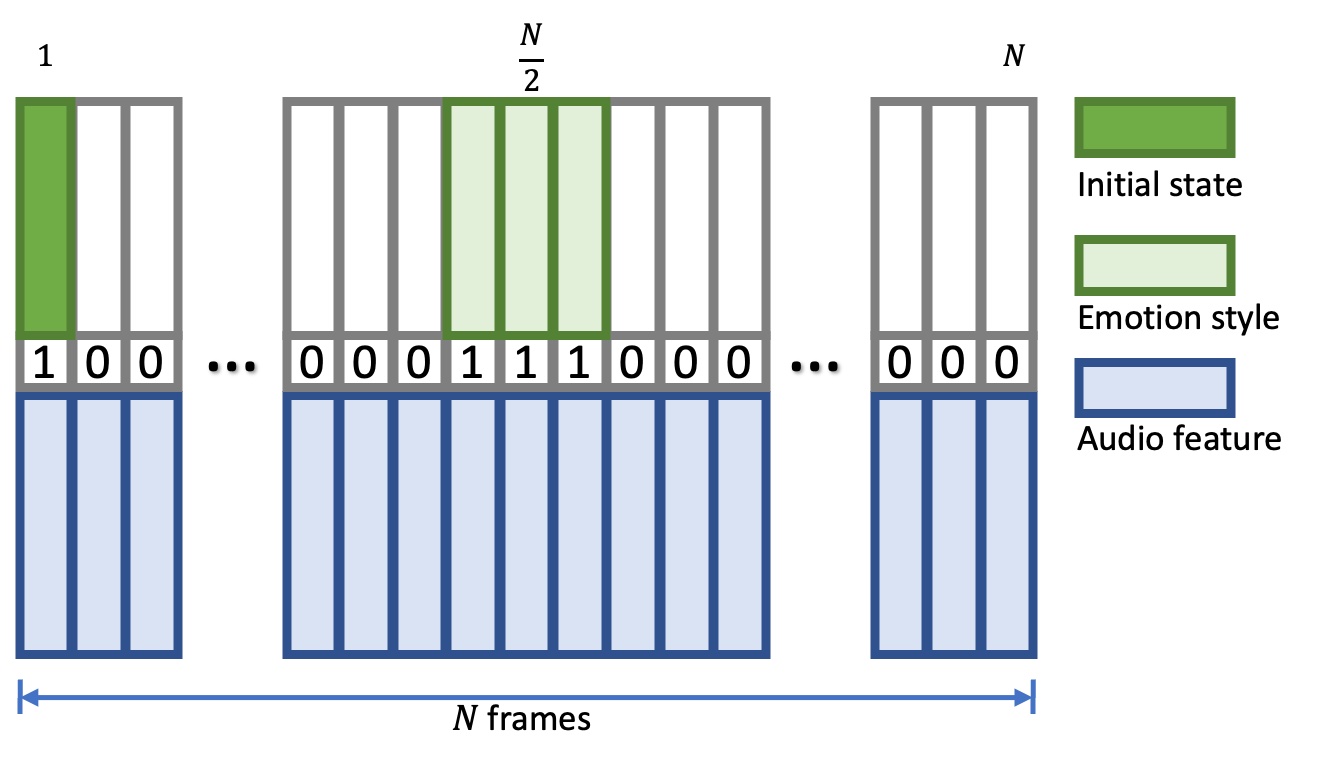}
\end{center}
\vspace{-3mm}
   \caption{Time position-aware embedding of conditions with N frames per sequence. The initial state corresponds to the first frame of the audio, and the middle three frames correspond to the emotion style. An additional bit is used to indicate the effective information. Finally, initial state and emotion style conditions are merged frame-by-frame with audio features.
   } 

\label{fig:embedding}
\end{figure}

\subsection{EmoDiff Module}
\label{sec:3.2}
Our goal is to generate 3D emotional expressions from audio. However, this task presents significant challenges that require innovative solutions. Firstly, mapping audio to expressions is a one-to-many problem, making it difficult to obtain dynamic and realistic expressions. Secondly, generating a sequence of 3D face expression parameters involves numerous issues, such as continuity and diversity.
To address these challenges, we propose the first diffusion-based model for generating 3D face expression sequences.

\noindent
\textbf{Forward diffusion and reverse process}.
We adopt the definition of denoising diffusion probabilistic models (DDPM)~\cite{ho2020denoising} as a forward noising process with latents $\{\bm{x}_t\}_{t=1}^T$ follow a Markov process $q(\bm{x}_t | \bm{x_0})$, where $\bm{x_0}$ is a data point sampled from a real data distribution $\bm{x_0} \sim q(\bm{x})$. The forward noising process can be defined as:
\begin{equation} \label{eq:1}
{
q(\bm{x}_{t} | \bm{x_0}) \sim \mathcal{N}(\sqrt{\bar{\alpha}_{t}}\bm{x_0}, (1 - \bar{\alpha}_{t})\bm{I}),
}
\end{equation}
where $\bar{\alpha}_{t} \in (0,1)$ follow a monotonically decreasing schedule, such that as $\bar{\alpha}_{t}$ approaches 0, $\bm{x}_{T}$ can be approximated as $\mathcal{N}(0,\bm{I})$. The data point $\bm{x_0}$ gradually loses its distinguishable features as step $t$ becomes larger, generating a sequence a sequence of latent variables $\bm{x}_1, \dots, \bm{x}_T$.

The reverse diffusion process estimates the joint distribution of $p_{\theta}(x_{0:T})$. The true sample could be generated from a noise input $\bm{x}_{T} \sim \mathcal{N}(0,\bm{I})$. To adapt to the conditional synthesis, we provide additional conditions $c$ to the model, including audio, initial state, and emotion style. The overview of our conditional reverse diffusion process is illustrated in Figure 3. The reverse process of each timestep can be updated as follows:
\begin{equation}
p_{\theta}(x_{t-1} \vert x_t, c) = \mathcal{N}(x_{t-1}; \mu_{\theta}(x_t, t, c), \beta_t I).
\label{eq:conditional}
\end{equation}
In the reverse diffusion process, given a Gaussian noise $x_T \sim \mathcal{N}(0, I)$, we use Eq. \ref{eq:conditional} to iteratively obtain the final results. Following~\cite{zhu2023taming}, we set the variances to $\beta_t I$ with untrained time-dependent constants. 

\noindent
\textbf{Training objective}.
Since $x_t$ is available as the input to the model, we predict the gaussian noise $\epsilon$ instead of $\mu$ at time step $t$:
\begin{align}
\mu_\theta(x_t, t, c)  = \frac{1}{\sqrt{\alpha_t}}\left( x_t - \frac{1-\alpha_t}{\sqrt{1-\bar\alpha_t}} \epsilon_\theta(x_t, t, c) \right),
\end{align}
where $\epsilon_\theta$ is a function approximator intended to predict $\epsilon$ from $x_t$.
We optimize $\theta$ with the 
following
objective, which works better by ignoring the weighting term introduced in Ho et al.~\cite{ho2020denoising}:

\begin{equation} \label{eq:Simloss}
\begin{split}
 \mathcal{L}_\text{simple}(\theta) =& \mathbb{E}_{t,\bm{x_0}, \bm{\epsilon}}\left[\| \bm{\epsilon} - \bm{\epsilon}_\theta(\bm{x}_t, t, \bm{c})\|_2^2\right]\\
 =& \mathbb{E}_{t,\bm{x_0}, \bm{\epsilon}}\left[\| \bm{\epsilon} - \bm{\epsilon}_\theta(\sqrt{\bar\alpha_t}\bm{x}_0 + \sqrt{1-\bar\alpha_t}\bm{\epsilon}, t, \bm{c})\|_2^2\right], 
\end{split}\end{equation}
where $t$ is uniformly sampled between $1$ and $T$.

\noindent
\textbf{Classifier-free guidance}.
We train our conditional diffusion model by applying classifier-free guidance~\cite{ho2022classifier}, which is widely used in recent diffusion-based works~\cite{tseng2022edge,zhu2023taming}. Specifically, the condition $\bm{c}$ gets discarded periodically at random by setting $\bm{c} = \phi$. The guided inference is then expressed as the weighted sum of unconditionally and conditionally generated samples:
\begin{align}
   \hat{\bm{\epsilon}}_\theta = (w+1)\bm{\epsilon}_\theta(\bm{x}_t, t,\bm{c}) - w(\bm{\epsilon}_{\theta}(\bm{x}_t, t,\phi),
\label{eq:epsmodified}
\end{align}
where $w$ is the scale parameter to trade off unconditionally and conditionally generated samples. The classifier-free guidance can achieve a good balance between quality and diversity.

\noindent
\textbf{Diffusion model}.
\label{sec:Model}
For the selection of the network architecture, we primarily considered two issues: 1) how to incorporate different modality data, such as audio and 3D expressions, and 2) how to generate the N-frame 3D face expression sequence different from images.
As shown in Fig.~\ref{fig:framework}, we choose to use the transformer structure, which fuses the representation of different modalities and captures the long-term dependency with a cross-attention mechanism following~\cite{zhu2023taming}. Please refer to the supplementary materials for detailed information on the network architecture.

\noindent
\textbf{Time-aware positional embedding}.
Generating facial expressions with temporal continuity requires consideration of two issues. 
Firstly, our training dataset only includes T-frame input data. However, during testing, we may require longer testing audio, so we need to consider the continuity between generated sequences. To address this issue, we  use the first frame of the generated sequence as the input feature condition and add it to the network, which constrains the initial state of the generated sequence.

Additionally, we need to guide the network to capture the style of emotion expression. To achieve this, we use three frames of expression as a representation of emotion style. To prevent the style from representing audio information, we randomly sample three frames of expression during training.
To incorporate the initial state condition and style information into the network, we use a time position-aware approach as illustrated in Fig.~\ref{fig:embedding} Specifically, we first create a matrix with the same length as the audio frames and set the first column of the matrix to the initial state value, a 50-dimensional vector with the last element set to 1. For style information, we select the middle three positions of the matrix and set them to the style values, with the last element also set to 1. We then combine this matrix with the audio feature frame by frame during training, which completes the setting of conditions.

\begin{table*}[!htbp] 
\centering
\caption{Comparison with state-of-the-art one-shot methods on MEAD and HDTF datasets. MakeitTalk and SadTalker maintain lip-sync and image quality without considering emotion. However, adding emotion complicates lip-sync and rendering. EAMM struggles with these challenges. Our method achieves both emotional expression and maintains lip-sync and image quality.}
\setlength{\tabcolsep}{2.0mm}{
\begin{tabular}{c|ccccc|ccccc}
\toprule  
&\multicolumn{5}{c|}{MEAD}&\multicolumn{5}{c}{HDTF} \\
\cmidrule(r){2-6}  \cmidrule(r){7-11}
Method & LPIPS$\downarrow$ & CPBD$\uparrow$ & F-LMD $\downarrow$ & LSE-D $\downarrow$ &LSE-C$\uparrow$ & LPIPS$\downarrow$  & CPBD$\uparrow$ & F-LMD $\downarrow$  &   LSE-D $\downarrow$  & LSE-C$\uparrow$ \\
\midrule
Ground Truth  & 0 & 0.316 & 0 & 7.420  & 7.486 &    0  & 0.303  &  0  & 7.413 & 7.487 \\
MakeitTalk   & 0.295 & 0.213 & 4.178 & 10.151 & 5.012 & 0.289 &  0.247 & 5.026 & 10.334  &  4.823 \\
SadTalker   & 0.189 & 0.256 & 3.960 & \textbf{9.634} & \textbf{6.095} & 0.195 & 0.269 & 4.006 & 9.958 &  5.050 \\
EAMM & 0.295 & 0.172 & 6.053 & 10.890 & 4.328 & 0.304 & 0.161  & 6.941 & 10.686  & 4.448  \\

\cmidrule(r){1-11}
\textbf{Ours}  & \textbf{0.169} & \textbf{0.299} & \textbf{3.845}  &   9.868   & 5.915   & \textbf{0.176} & \textbf{0.280} & \textbf{3.948}   & \textbf{9.233} & \textbf{5.263}  \\
\bottomrule 
\end{tabular}
}
\label{table:quantitive_evaluation}
\end{table*}

\begin{algorithm}[t]
  \caption{\textbf{Long-term Dynamic Sampling}} \label{alg:sampling}
  \small
\begin{algorithmic}[1]
     \State Trained diffusion model $M$,  Input audio $A$, Emotion Style $S$, Frame length $N$.
     \State $L_A=length(A)$
     \State $L_{exp}=length(S[0])$
     \State Output $ o = zeros(L, L_{exp}) $
     \State Condition $ c = zeros(N, N_{exp}+1)$
     \State $ c[:,0,:-1] = S[random(1),:]$
     \State $ c[:,N/2-1:N/2+2,:-1] = S[random(3),:]$
     \State $ c[:,0,-1] = 1 $ 
     \State $c[:,N/2-1:N/2+2,-1] = 1 $
    \For{$i=0::N-1::L_A$}
        \State temp$=$M.sampling$(cat[c,A[i:i+N]])$
        \State $ c[:,0,:-1] = temp[-1]$
        \State $ c[:,N/2-1:N/2+2,:-1] = S[random(3),:]$
        \State $o[i:i+N]=temp$
    \EndFor
    \State \textbf{return} $\bm{o}$
  \end{algorithmic}
\end{algorithm}
\noindent
\textbf{Long-term and dynamic sampling}.
During testing, we first select an emotional clip of a character from the dataset, such as "M003 Angry clip001", along with input audio. As shown in Alg. \ref{alg:sampling}, to ensure continuity between sequences, we randomly select one frame as the initial state, and subsequently use the last frame of the previous sequence as the initial state for the next sequence. This ensures the continuity of long sequences. To introduce diversity in each sequence generation, we randomly select 3 frames as the style each time, which allows for the generation of dynamic facial expressions within the overall sequence.

\subsection{Lip Refinement}

After obtaining dynamic emotional expressions denoted as $\bm{x}_{0}$ from the diffusion model, we observed an unintended consequence in which the diffusion network inadvertently reduced the influence of audio, resulting in a noticeable misalignment between the audio and mouth shape. This phenomenon is likely attributed to the diffusion network's inherent inclination toward generating realistic sequences, which, in turn, diminishes the impact of the audio. To rectify this issue, we introduce a lip-sync refinement network that utilizes the same audio and emotional data to recalibrate and generate refined mouth parameters. Our Lip-sync network incorporates an LSTM structure as the audio encoder and a CNN structure as the emotion encoder. This design effectively generates mouth-related parameters that closely align with the input audio and emotional reference style. For a comprehensive understanding of our lip refinement network, we direct readers to the supplementary materials.

Subsequently, we employ these refined facial parameters and the generated head poses to animate a 3D blend shape rig. Utilizing GPU rendering, we obtain corresponding 3D rendered avatar images denoted as $I_{t,t\in \{0,..,N\}}$. Following this, we employ a video-to-video approach to generate talking face videos for arbitrary characters. Our ablation studies demonstrated a notable improvement in the synchronization between mouth movements and audio upon the implementation of lip refinement.

\subsection{Face Neural Rendering}
\label{sec:3.3}
Upon acquiring synthesized images $I_{t,t\in \{0,..,N\}}$ from an external GPU renderer, we employ motion transfer techniques to achieve a realistic talking head effect for different subjects. Specifically, we utilize the Face-Vid2Vid method proposed by Wang et al~\cite{wang2021facevid2vid} as the fundamental neural rendering pipeline $\mathcal{R}$. Furthermore, we conduct a fine-tuning process on the model using carefully selected high-resolution expressive talking videos from TalkHead-1HK dataset~\cite{wang2021facevid2vid}, aiming to enhance both expressiveness and rendering quality. In addition to fine-tuning, we augment the final image resolution to 512x512 using the face super-resolution method outlined in~\cite{chan2021glean}. To ensure effective identity preservation throughout the process, we implement the relative mode technique developed by Siarohin et al. ~\cite{Siarohin_2019_NeurIPS} for neural motion transfer. Specifically, we first render a reference frame $I_n$ with neural expression and then apply the relative motion $\mathcal{M}_{I_n \rightarrow I_t}$, which represents the transformation between talking frames and the neural frame, onto the source image $T$. Consequently, ultimate rendered outputs $\mathcal{R}(T, \mathcal{M}_{I_n \rightarrow I_t})$ is generated.

\begin{table}[t]
\begin{center}
\caption{Comparisons of state-of-the-art methods and our proposed method. SadTalker and MakeItTalk do not generate emotional speech. EAMM produces emotional videos but loses identity and dynamic facial expressions. EVP is a video-based method that generates emotional speech but lacks emotion dynamics. In contrast, DreamTalk offers dynamic emotional expression with generated eye blinks and identity preservation.}
\label{tab:listcompare} 
\begin{tabular}{cccccc}
\toprule
 \multirow{2}{*}{Methods} & Emotional & Generated & Identity  \\
& talking & eye blinks & preservation  \\
\midrule
 MakeItTalk &  \xmark & \xmark & \cmark  \\
SadTalker  &\xmark & \cmark & \cmark \\
EAMM & \cmark & \xmark & \xmark  \\
EVP & \cmark & \xmark & \cmark  \\
Ours & \cmark & \cmark & \cmark   \\

\bottomrule
\end{tabular}
\end{center}
\vspace{-5mm}
\end{table}

\begin{figure*}[t]
\begin{center}
   \includegraphics[width=\linewidth]{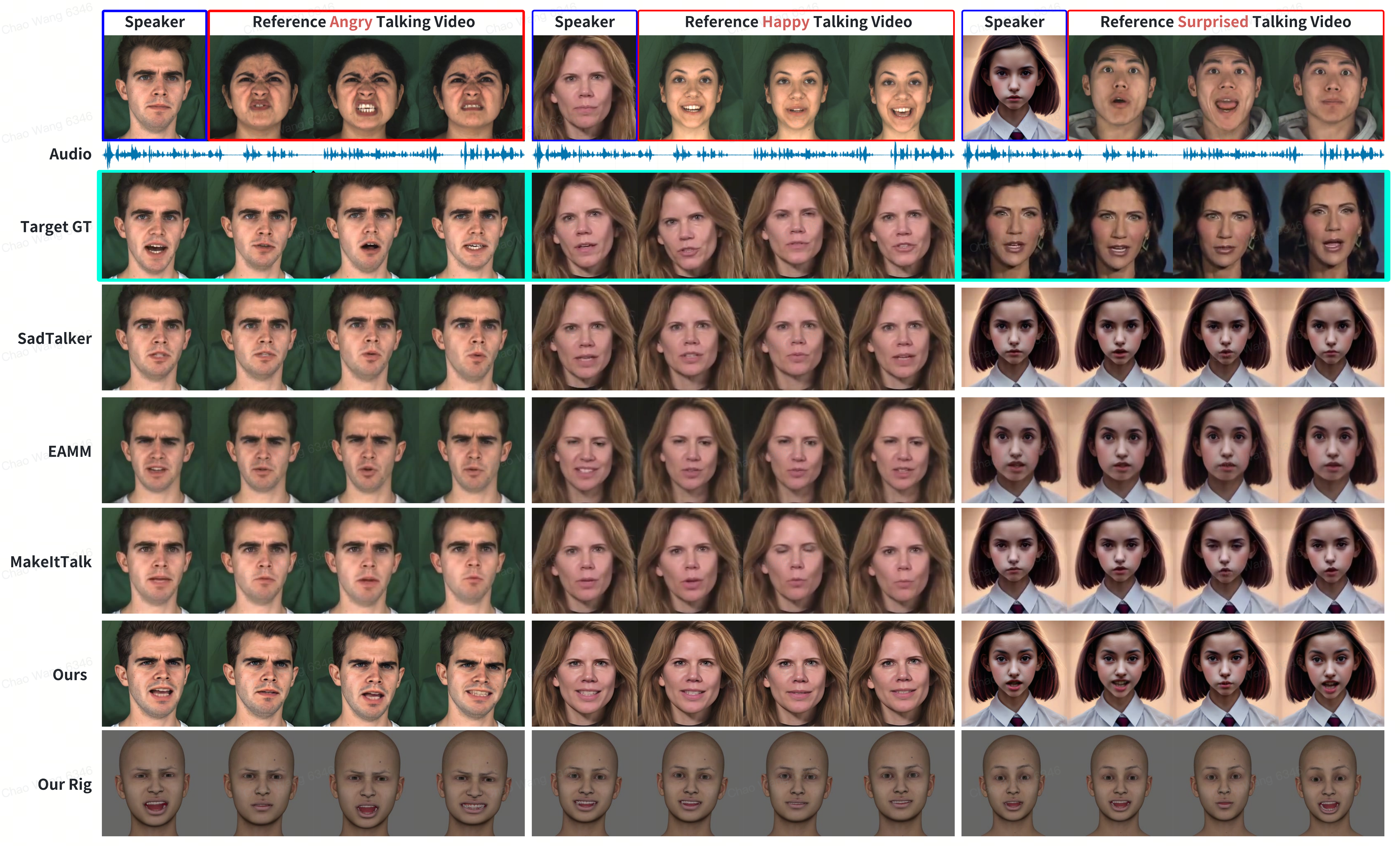}
\end{center}
\vspace{-1mm}
   \caption{
Comparison of state-of-the-art models with our approach: In the first two comparisons, we conduct evaluations on the MEAD and HDTF datasets, respectively. For the third comparison, we utilize one AIGC-generated face. We also visualize our rig model results as intermediate representations. Our method consistently yields significantly superior results in terms of emotional expression, lip synchronization, identity preservation, and image quality. Please refer to our supplementary video for better comparison.
   } 
\label{fig:main_result}
\end{figure*}

\section{Experiments}

\subsection{Implementation Details}
All experiments were conducted on a single V100 GPU utilizing the Adam optimizer~\cite{kingma2014adam}. The frame rate for all training datasets was set at 25 FPS. In the EmoDiff Module, our training primarily leveraged two datasets: the MEAD emotion dataset ~\cite{wang2020mead} and the HDTF multi-character dataset~\cite{zhang2021flow}. Each sequence generated during training consisted of a fixed length of 32 frames. We trained for a total of 1000 epochs with a batch size of 64 and a learning rate of 0.0004.
For the Lip refinement model, we employed a sliding window of size $T=8$ to extract training samples of audio features. The training process encompassed 50 epochs with a batch size of 32 and a learning rate of 0.0001. 

\subsection{Comparison with State-of-the-Arts}
\begin{figure}[t]
\begin{center}
\vspace{-3mm}
\includegraphics[width=1.0\linewidth]{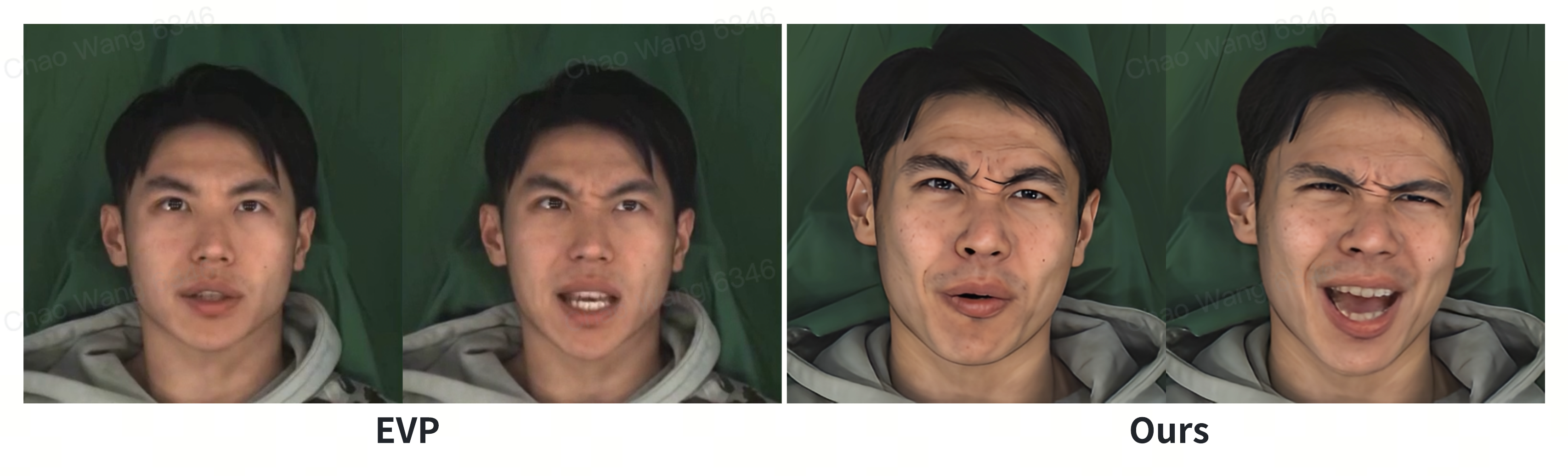}
\end{center}
\vspace{-3mm}
   \caption{Comparison with the video sequences provided by EVP on emotion "angry". Since EVP requires separate training for each video, we cannot test it with arbitrary characters.}
\label{fig:evp}
\vspace{-5mm}
\end{figure}

In~\tabref{tab:listcompare}, we offer an intuitive comparison of various methods' capabilities. It's clear that SadTalker and MakeItTalk lack the ability to generate emotional speech. Although EAMM can produce emotional videos, it does so at the expense of maintaining identity and fails to incorporate dynamic facial expressions, such as eye blinking. While EVP can generate emotional speech (Fig. \ref{fig:evp}) , it also sacrifices the dynamism of emotions and cannot drive from a single image, limiting its applicability.
In contrast, DreamTalk not only guarantees dynamic emotional expressions, such as generating high-frequency expressions like eye blinks, but also delivers high-quality videos with accurate lip-sync.

\begin{figure}[t]
\begin{center}
\includegraphics[width=1.0\linewidth]{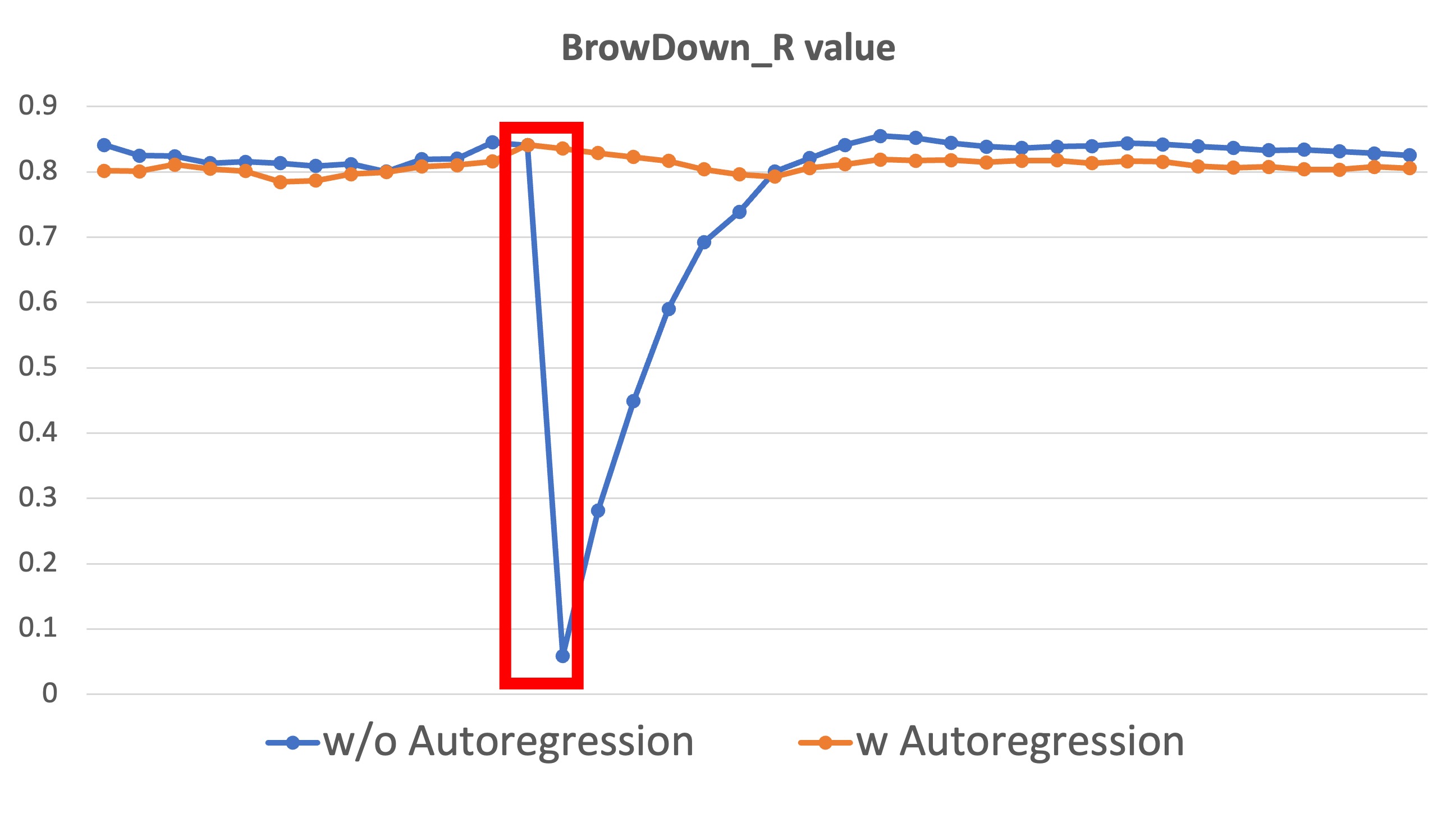}
\end{center}
\vspace{-5mm}
   \caption{Here we evaluate the control parameters of Brow Down on the right side. There will be a big jump between generated sequences, such as the results in the red box. By employing autoregressive feedback to generate long sequences, we can ensure continuous generation results.}
\label{fig:ablation1}
\vspace{-3mm}
\end{figure}

\subsubsection{Qualitative Evaluation}
As shown in~\figref{fig:main_result}, we compare our work with three state-of-the-art methods. MakeItTalk and SadTalker can't generate desired emotions from a single image. MakeItTalk's 2D approach lowers image quality, while SadTalker provides only neutral conversation. EAMM generates emotional speech but sacrifices emotion dynamics and video quality. Our method excels in emotional expression, lip-sync, identity preservation, and image quality.

\subsubsection{Quantitative Evaluation}

Following established standards in the field, we utilize metrics for lip sync and image quality in our comparative analyses. 
For assessing the synchronization between lip movements and input audio, we employ SyncNet~\cite{chung2016out}, which measures he distance
score (LSE-D) and confidence score (LSE-C) to evaluate lip-sync precision. We also employ the Landmark Distance on the entire face (F-LMD) for a comprehensive evaluation of facial expression accuracy. 
For image quality, we assess image quality using learned perceptual image patch similarity (LPIPS) and cumulative probability blur detection (CPBD). 


Our model exhibits enhanced performance over current leading methods, as demonstrated in~\tabref{table:quantitive_evaluation}.
Comparing state-of-the-art methods on MEAD and HDTF datasets, MakeItTalk and SadTalker maintain lip-sync and image quality without considering emotion. However, adding emotion complicates lip-sync and rendering. EAMM faces challenges in achieving both emotion and maintaining lip-sync and image quality. Our method successfully balances emotional expression with lip-sync and image quality.


\subsection{Ablation Study}
Our ablation study primarily investigates three key questions, including whether to employ an auto-regressive generation approach based on the initial state, whether to utilize a style encoding method with positional embedding, and whether to incorporate lip refinement for lip-sync results.

\noindent \textbf{Autoregressive generation}
Since our goal is to generate indefinitely long sequences, we require an autoregressive method to achieve continuous talking results. In this approach, we use the last state of the previous sequence as a condition to generate the next sequence, thereby ensuring continuity between sequences. As illustrated in~\figref{fig:ablation1}, where we depict the Brow Down value sequence, it is evident that without the autoregressive constraint, there will be a big jump between generated sequences.

\noindent \textbf{Emotional style embedding}
Unlike the approach used in EAMM, our emotional style employs positional embedding with diffusion model. Since EAMM applies the same style for each frame, which diminishes the dynamism of the emotional style, resulting in overly smooth emotions that fail to capture high-frequency information, such as eye blinking. Here, we compare the use of a uniform emotional code against our method of positional embedding. It is observed that our diffusion method combined with positional embedding effectively captures high-frequency information. Refer to our supplementary video for comparison.

\noindent \textbf{Lip refinement}
While the diffusion network aids in generating dynamic emotions, we observed that it struggles to produce sequences that fully align with the audio. Hence, we employed a Lip Refinement Model to further optimize lip motion based on the audio. As shown in~\tabref{tab:ablation1}, we measured the results using SyncNet and found that the use of Lip Refinement leads to more synchronized lip motion.

\begin{table}[t]
\caption{Ablation for lip refinement. Without lip refinement, the quality of the generated mouth region cannot be guaranteed. With lip refinement, the synchronization between the generated mouth and the audio improves, and mouth movements becomes closer to the ground truth (GT).}
\vspace{-5mm}
\begin{center}
\setlength{\tabcolsep}{1.5mm}
\renewcommand{\arraystretch}{1.1}
\begin{tabular}{l|ccc }
\toprule
 Method  & F-LMD $\downarrow$ & LSE-D $\downarrow$ &LSE-C$\uparrow$ \\
\midrule
w/o Lip refinement & 4.167 &  10.583 &  4.338\\
Ours & \textbf{3.927} &  \textbf{9.648} &  \textbf{5.507}\\
\bottomrule
\end{tabular}
\end{center}

\label{tab:ablation1} 
\vspace{-5mm}
\end{table}

\begin{figure}[t]
\begin{center}
\includegraphics[width=0.9\linewidth]{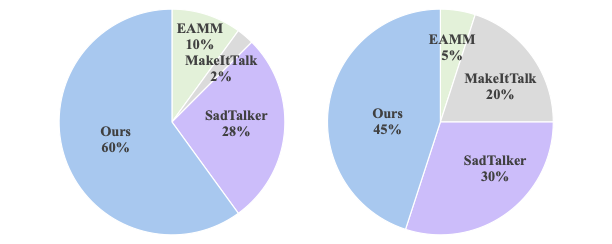}
\end{center}
\vspace{-4mm}
   \caption{User study results show ratings for emotion preservation (on the left) and overall quality (on the right).}
\label{fig:Userstudy}
\vspace{-4mm}
\end{figure}

\subsection{User Study}
Due to the subjective nature of emotion, quantitative evaluation is challenging. Therefore, we employed a subjective assessment method, involving 20 users who compared the results of different speech generation techniques. We provided reference images and emotional information for evaluation. 
The evaluation results depicted in Figure \ref{fig:Userstudy} demonstrate
that while EAMM can generate emotion, it comes at the expense of video quality, leading to lower user ratings. Makeitalk and SadTalker, while lacking in emotion generation, achieved better overall quality than EAMM. 
Our method, on the other hand, successfully maintains emotional intensity while attaining high-quality generation.
\section{Conclusion}
In this paper, we present DREAM-Talk, an innovative framework designed for the generation of emotionally expressive talking faces with precise lip synchronization. Our two-stage approach, comprising the EmoDiff Module and Lip Refinement, effectively captures emotional nuances and ensures accurate lip-syncing. Leveraging an emotion-conditioned diffusion model and a lip refinement network, our method outperforms existing techniques. Our results demonstrate improved facial emotional expressiveness while maintaining high video quality. 
DREAM-Talk represents a significant leap forward in the domain of emotional talking face generation, enabling the creation of realistic and emotionally engaging digital human representations across a wide range of applications.
{
    \small
    \bibliographystyle{ieeenat_fullname}
    \bibliography{main}
}

\end{document}